%% file: lrec-coling2024-example.tex
\documentclass[10pt, a4paper]{article}

\usepackage{lrec-coling2024} 

\usepackage{times}
\usepackage{latexsym}

\usepackage[T1]{fontenc}

\usepackage[utf8]{inputenc}

\usepackage{microtype}

\usepackage{inconsolata}

\usepackage{graphicx} 
\usepackage{parskip}  

\usepackage{multirow}
\usepackage{amssymb}
\usepackage{subfig}
\usepackage{comment}

\title{Semantic Map-based Generation of Navigation Instructions}

\name{
Chengzu Li$^{1,2}$\sthanks{\ \ Work done on internship at Toshiba Europe Limited.}, 
Chao Zhang$^{2}$, 
Simone Teufel$^{1}$, \\
\textbf{\large Rama Sanand Doddipatla$^{2}$, 
Svetlana Stoyanchev$^{2}$}
} 

\address{
$^{1}$University of Cambridge,
$^{2}$Toshiba Europe Limited \\
\{cl917, sht25\}@cam.ac.uk \\
\{chao.zhang, rama.doddipatla, svetlana.stoyanchev\}@toshiba.eu
}

\input{sections/abstract}

\begin{document}

\maketitleabstract

\input{sections/introduction}

\input{sections/task_definition}

\input{sections/methods}

\input{sections/results_and_discussion}

\input{sections/conclusion}

\input{sections/limitations}

\input{sections/data_code_availability}

\nocite{*}
\section*{Bibliographical References}\label{sec:reference}

\bibliographystyle{lrec-coling2024-natbib}
\bibliography{lrec-coling2024-example}

\appendix
\input{appendix}

\end{document}

%% file: sections/abstract.tex
\abstract{
We are interested in the generation of navigation instructions, either in their own right or as training material for robotic navigation task. 
In this paper, we propose a new approach to navigation instruction generation by framing the problem
as an image captioning task using semantic maps as visual input.  
Conventional approaches 
employ a sequence of panorama images to generate navigation instructions. 
Semantic maps abstract away from visual details and fuse the information in multiple panorama images into a single top-down representation, thereby reducing computational complexity to process the input.
We  present a benchmark dataset for instruction generation using semantic maps, propose an initial model  and ask human subjects to manually assess the quality of generated instructions.  
Our initial investigations show promise in using semantic maps for instruction generation instead of a sequence of panorama images, but there is vast scope for improvement. 
We release the code for data preparation and model training at \url{https://github.com/chengzu-li/VLGen}. 
 \\ \newline \Keywords{semantic map, navigation instruction generation, Room2Room} 
 }

%% file: sections/introduction.tex
\section{Introduction}
Vision and Language Navigation (VLN) is a task that involves an agent navigating in a physical environment in response to natural language instructions \citep{wu2021visual}.
The data annotation for the VLN task is time-consuming and costly to scale up, and
the development of models that address the task is severely limited by the availability of training data \citep{gu-etal-2022-vision}. 
Navigation instruction generation (VL-GEN) is the reverse of the VLN task in that it generates natural language instructions for a path in the virtual (or physical) environment, which is helpful for interactions with users and explainability. 
Previous work has also demonstrated the effectiveness of VL-GEN in improving the performance of VLN systems such as the Speaker-Follower model \citep{fried2018speaker} and Env Drop  \citep{tan-etal-2019-learning}. 
This paper explores the VL-GEN task of generating navigation instruction framing it as an image captioning task. 

VL-GEN requires the model to generate language instruction in the context of the physical environment, grounding objects references and action instructions to the given space. 
Previous studies use photo-realistic RGB panoramic images as the visual input; they frame VL-GEN as the end-to-end task of generating text from a sequence of photo-realistic RGB images~\citep{fried2018speaker, tan-etal-2019-learning, wang2022less}.
While \citet{zhao2021evaluation} report that the overall quality of instructions generated with end-to-end models is only slightly better than that of template-based generation, the application of object grounding to the panoramic images achieves a better result \cite{wang2022less}. 

The existing approach to this task has two shortcomings. 
From the perspective of representation, using  panoramic images is resource-intensive as it requires processing of multiple image inputs corresponding to different points on the path. 
Second, panoramic images contain many details that are irrelevant for the task.  
The model has to learn to interpret the environments from RGB panoramas, such as object recognition, and generate instructions at the same time. 
As it is natural for humans to understand navigation instructions from a top-down map (as in Google Maps) \cite{paz-argaman-etal-2024-go}, we propose to separate the VL-GEN task into two steps: 1) environment interpretation, which is addressed by semantic SLAM in physical robotic systems \cite{chaplot2020learning}, and 2) spatial reasoning. 
In this paper, we focus on the second step and explore the feasibility of using top-down semantic map for VL-GEN. 

\input{figs/illustration/data_example}
Our research question is whether it is feasible to use the top-down semantic map (a single RGB image) as our main source of information.
We also explore which other data sources, in addition to the semantic map, can further improve performance. 
To address this question, we formalize the VL-GEN task as image captioning with the input of a semantic map with the path (see Figure \ref{fig:data_example}). 
We extract the images of top-down maps  from the Habitat simulator \citep{savva2019habitat} based on Room-to-Room dataset \citep{Anderson_2018_CVPR} and VLN-CE \citep{krantz2020beyond}. 
Our key contributions and findings include the following:

\begin{itemize}
    \item We extend the R2R dataset with semantic maps, providing a new benchmark dataset and a baseline that demonstrates the feasibility of using semantic maps for VL-GEN task. 
    \item We demonstrate experimentally with both automatic and human evaluations that including additional information (namely, region, action, and prompt) leads to more accurate and robust navigation instructions than using only semantic maps. 
    \item We also conduct an intrinsic human evaluation of the quality of the generated instructions with fine-grained error analysis.
\end{itemize}

%% file: figs/illustration/data_example.tex
\begin{figure}[t!]
    \centering
    \includegraphics[width=1\columnwidth]{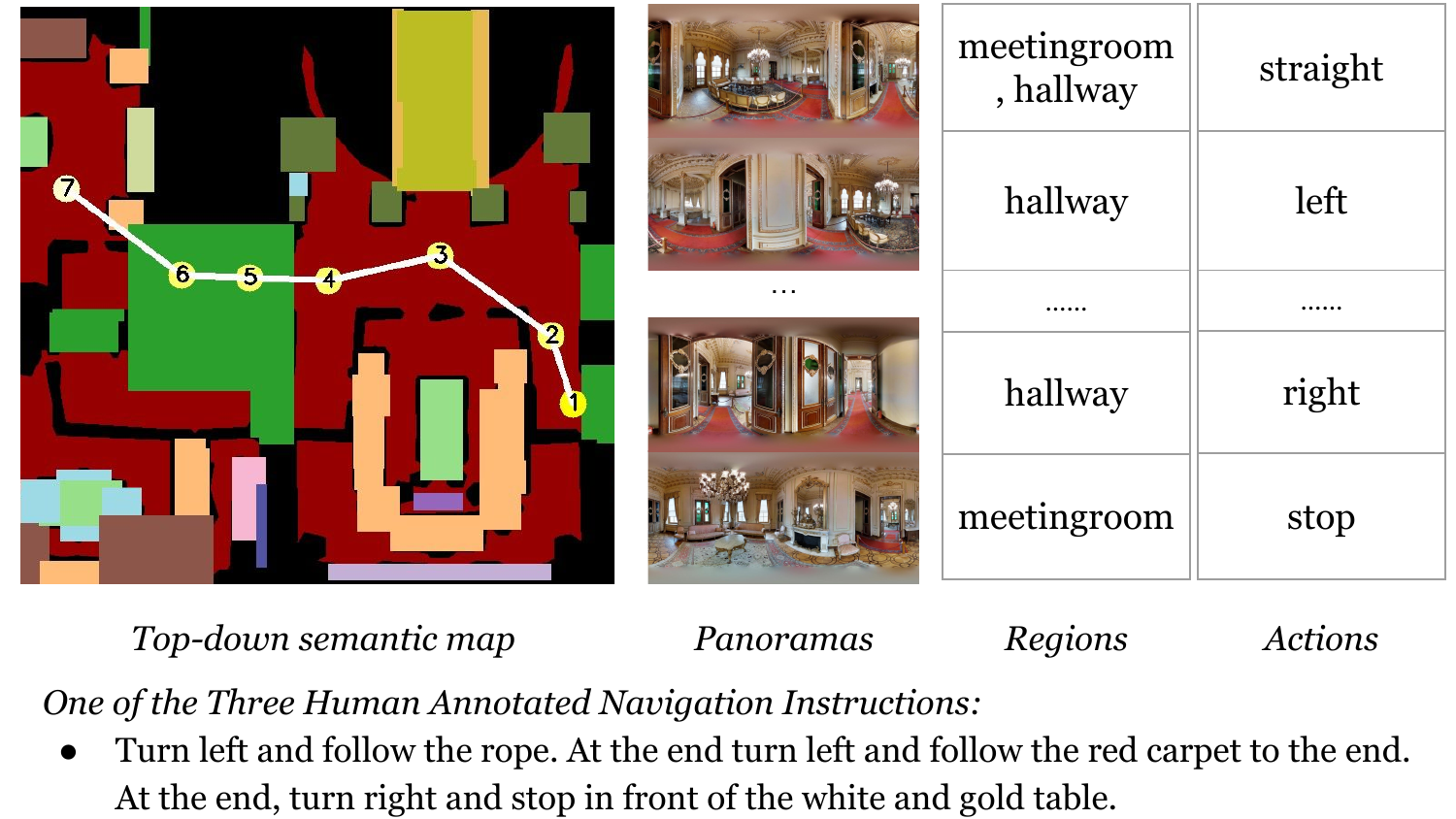}
\caption{
An example navigation scenario from our new dataset for instruction generation, with the navigation path overlayed on the semantic map.
} 
\label{fig:data_example}
\end{figure}

%% file: sections/task_definition.tex
\section{Task Definition and Data}

A semantic map $M_s$ is a top-down view of the scene $s$, which contains a path $P=\{p_1,...,p_K\}$, represented as a sequence of points connected by a line, and a set of $N$ objects $O=\{o_1,...o_N\}$. 

In light of the success of image captioning models~\cite{li2022blip, wang2022ofa}, we frame the VL-GEN task as image captioning task.
Given a semantic map $M_s$, the task is to generate a natural language description $D_P$ that describes the path $P$ shown.
Our task description replaces the photo-realistic RGB images used previously, with a semantic map. 
The processing of RGB images is resource-intensive, while our task definition has the advantage of abstracting away from the object recognition task, concentrating on the instruction generation task instead. 

\input{tables/dataset_info}

We also experiment with providing the model with additional features of the navigation path beyond the semantic maps alone, including actions, names of regions, and panoramic images. 
There is a fixed set of action types (\textsc{left, right, straight, stop}), which are determined heuristically from the path shape at each navigation point.
For each navigation point, we use the name of its associated region (e.g., hallway, meeting room). 
We do not think that panoramic images constitute ideal input to the system, 
but it is possible that they may provide additional visual information not shown in the map. 
Therefore, we also conduct experiments with panoramic images as part of the input information to the model. 

We extract semantic maps, region and action information from the Habitat \citep{savva2019habitat, krantz2020beyond} simulation environment. 
In a deployed robot, it may be obtained with a semantic SLAM component~\cite{chaplot2020learning}.
Each object type on the map is represented in a unique color. 
We adopt the navigation paths and human annotations from the R2R dataset \citep{Anderson_2018_CVPR}. 
Panoramic images in RGB are obtained from the Matterport3D simulator \citep{chang2017matterport3d} at each discrete navigation point.
An example of the new dataset derived from R2R,  including a semantic map with a path, language instruction, panorama images, actions, and region names, is shown in Figure~\ref{fig:data_example}. 

Statistics about the semantic maps are presented in Table \ref{tab:dataset_info}. 
The data splits we use are inherited from the original R2R dataset. 
The difference between seen validation set and the unseen validation set in R2R is whether the room environment is included in the train set.\footnote{Further details on the dataset are presented in Appendix \ref{app:data extraction}.}

%% file: tables/dataset_info.tex
\begin{table}[]
\centering
\renewcommand\arraystretch{1.3}
\resizebox{\columnwidth}{!}{%
\begin{tabular}{ccccc}
\hline
split      & size  & Avg. \# points & Avg. \# regions & Avg. \# objects  \\ \hline
train      & 10623 & 5.95           & 3.26            & 22.64                              \\
val seen   & 768   & 6.07           & 3.3             & 22.36                               \\
val unseen & 1839  & 5.87           & 3.11            & 22.13                                \\ \hline
\end{tabular}%
}
\caption{Statistics of extracted semantic maps. Avg. \# region: average number of distinct regions along the path. Avg. \# object: average number of object types in the semantic map. }
\label{tab:dataset_info}
\end{table}

%% file: sections/methods.tex
\section{Method}
Motivated by the success of the multimodal pre-trained models, we construct a multimodal text generation model using BLIP\footnote{The implementation is based on the Huggingface transformers library \citep{wolf2019huggingface}: \href{https://huggingface.co/Salesforce/blip-image-captioning-base}{Salesforce/blip-image-captioning-base}} \citep{li2022blip}.
Figure~\ref{fig:overall_model} illustrates the architecture of the proposed model with modules that process different inputs; these will be described in Section~\ref{sec:input}.  
In Section~\ref{sec:augmentation}, we describe the augmentations applied to the BLIP model in our experiments.

\input{figs/illustration/model_image}

\subsection{Model Input}
\label{sec:input}

\paragraph{Top-down semantic map \textit{(TD)}}
The semantic map forms the main input used in all experiments. 
It is encoded by the image encoder in the BLIP model. 
We first resize the image by nearest sampling to $384 \times 384$ and then feed it to the vision transformer with patch size 16.

\paragraph{Regions \textit{(Reg)} and actions \textit{(Act)}}
Region names and actions are frequently mentioned in human navigation instructions. 
To give the model information about the relevant region names, we represent them as a sequence of strings for each navigation point.
We use a text encoder from the pre-trained BLIP model to represent the region names. 
The region embedding for each point is obtained by applying a mean pooling operation to the word embeddings. 
For actions, we apply an embedding layer to the discrete action values and get action embeddings in the same dimension as the region embedding. 
We add the region and the action embeddings together 
at each point and use a 3-layer LSTM model to embed the sequential information along the navigation path. 

\paragraph{Panoramic images \textit{(Pano)}}
Based on our analysis, visual object properties such as color and shape are mentioned in more than 25\% of human instructions. 
As semantic maps only include object types but not the properties of visual objects, we augment the model input with panoramic images.
This might enable the model to learn the visual properties mentioned in the instructions. 
We initialize the image encoder based on the pre-trained image encoder in BLIP model. 
We freeze its parameters during training because the model is pre-trained on photo-realistic images, which we believe endows the model with capabilities of recognizing panoramic images in our case. 
In order to increase the flexibility of the visual embedding, we apply an additional MLP with two linear layers on top of the panoramic vision encoder. 
Following the methods in the video captioning task \citep{tang2021clip4caption, luo2022clip4clip}, we treat the panoramas as discrete frames and use the mean average of all panoramic embeddings to represent the panorama information of the navigation path.

Finally, the embedded input representations are added together to form the input to the decoder that outputs natural language instructions.

\subsection{Model Augmentation}
\label{sec:augmentation}
\paragraph{Multimodal alignment with contrastive loss}
Contrastive learning is an effective method used in self-supervised learning for visual representation learning \cite{radford2021clip, li2022blip} and multimodal pre-training in BLIP \citep{li2022blip}. 
We investigate the effectiveness of introducing contrastive training for navigation instruction generation task as an auxiliary loss.
We define the positive examples $P^{+}(C_{gt}, I_{gt})$ as pairs of the combined input embedding and the instruction embedding.
The negative examples $P^{-}(C_{gt}, I_{rnd})$ consist of the pairs of the input embedding and the embedding of a randomly sampled instruction. 
Following CLIP \citep{radford2021clip}, we multiply the multimodal input matrix $E_{input}$ and textual instruction matrix $E_{text}$ to obtain the predicted compatible matrix $C_{pred}$ between inputs and labels and then compute the CrossEntropy loss on $C_{pred}$ with the ground-truth correspondence $C_{gt}$. 

\paragraph{Augmentation and grounding with prompt}

The prompting of LLMs has demonstrated its effectiveness across various domains in previous works \citep{li2021prefix, liu2021p, tang2022context, keicher2022few, v2p}. 
We generate the prompt from a template, which describes the nearby objects and regions, such as \textit{Starting from the dark yellow point near sofa cushion in the living room region}.
We tune the model with prompting and feed the prompt template to the decoder during inference. 
We argue that prompting can benefit the generation task in two ways. 
First, it can help visual-language grounding because the prompting template describes nearby landmarks and regions.
Second, at inference time, the instructions that are generated are conditioned on the prompt template in an auto-regressive way, resulting in more controllable generation in VL-GEN task.

%% file: figs/illustration/model_image.tex
\begin{figure}[t!]
    \centering
    \includegraphics[width=1\columnwidth]{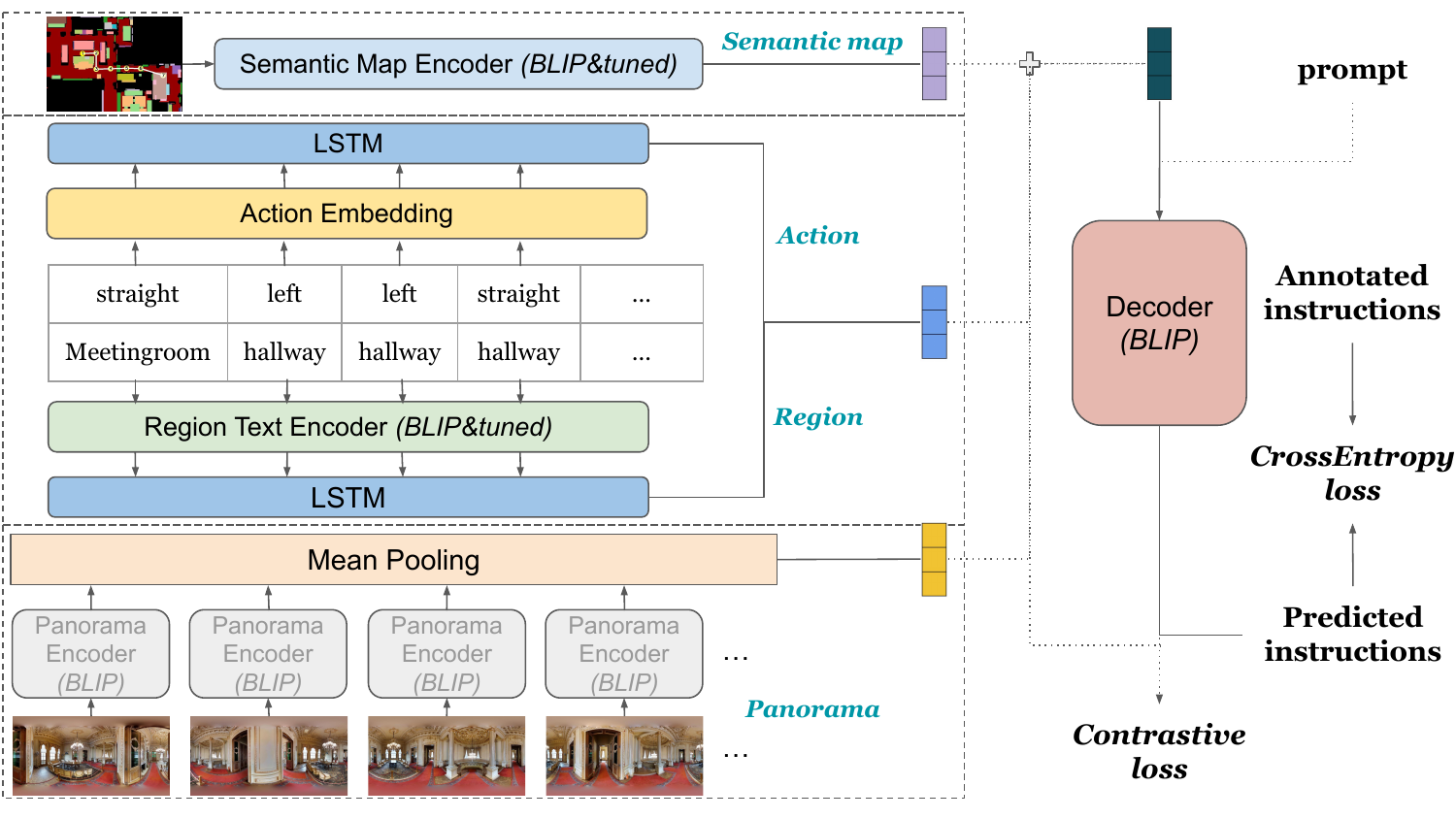}
    \caption{Illustration of the overall model architecture. Text input is encoded with pretrained BLIP text encoder and LSTM, and image input is encoded with the pretrained BLIP encoder. Modules shown in the same color share the weights. The weights of the panorama encoder are fixed. }
    \label{fig:overall_model}
\end{figure}

%% file: sections/results_and_discussion.tex
\section{Experiments}

\input{figs/results_bar}
We perform two evaluations over experiments: an automatic evaluation according to performance on the task (extrinsic) and a human evaluation of the quality of the instructions (intrinsic). 
These evaluations can tell us about the influence of region, actions, prompting, and contrastive loss on the quality of the instructions both quantitatively and qualitatively.

\subsection{Experimental setup}
We train the model using the train split of the R2R dataset and evaluate it both on validation seen and unseen sets. 
We use the BLIP-\textit{base} model for experiments.
We setup the baselines with different combinations of the input: 1) top-down semantic map (TD) 2) + regions (Reg) and actions (Act); 3) + panoramic images (Pano). We also experiment with contrastive loss and prompting, making 9 system variants for experiments in total. 

In the intrinsic human evaluation, we use a Latin Square design of size 5. 
We therefore compare only a subset of the above system variants with different combinations of input (\textit{TD}, \textit{TD+Reg+Act} and \textit{TD+Reg+Act+Pano}), and prompting and contrastive loss on \textit{TD+Reg+Act}.

\subsection{Human Participants and Procedure}

For the human experiment, we recruit 5 evaluators who have never contributed to or been involved in the project before under the consent from the Ethics Committee. The evaluation workload for each participant is designed to be within 30 minutes for them to concentrate on the task. We also provide two specific illustration examples about the evaluation task for the human participants.
The evaluation materials consist of 15 navigation paths in the unseen environments, randomly sampled. 
The experiment is performed online using an evaluation interface. 
The participants are shown the semantic map with the path as well as panorama images.
They are asked to assign a score from 0 (worst) to 10 (best) based on the quality of the instruction candidates generated by different systems. 

\subsection{Automatic Evaluation Metrics}

In the automatic evaluation, we compare the performance of 9 system variants based on an automatic metric SPICE (Semantic Propositional Image Caption Evaluation) \citep{spice2016}, following~\citet{zhao2021evaluation}.
SPICE is a metric used to evaluate the quality of image captions, focusing on the semantic content of captions. 
It identifies semantic propositions within the parse trees and compares the semantic propositions from the generated caption with those from the reference captions.

When comparing different systems, we use the two-sided permutation test to see if the arithmetic means of the two systems' performances are equal. 
If the p-value is larger than 0.05, we consider the performance of the two systems to be not significantly different. 

\subsection{Evaluation Results}
Table \ref{tab:eval_results} shows the SPICE and human evaluation scores in seen and unseen environments. 
As expected, the models perform better in seen than in unseen setting by 3.88 in SPICE score on average across all 9 systems. 
For both settings, we observe that using region and action information with the prompt improves the model's performance with $p\leq0.05$, while contrastive learning does not seem to help. 
Adding panoramic images tends to improve the performance, but not significantly ($p\geq0.1$).  
When comparing with previous methods in SPICE score, our systems (17.84/22.14) perform on par or even achieve higher SPICE scores than Speaker Fol. \cite{fried2018speaker} (17.0/18.7) and EnvDrop \cite{tan-etal-2019-learning} (18.1/20.2) on unseen/seen settings.

In the results for the human evaluation, shown in Table~\ref{tab:eval_results}, we observe that using the semantic map as the only input results in the lowest average score across all systems (3.42). 
This repeats the observations from the automatic evaluation.
Using regions, actions, and panoramas achieves the highest rating (4.36) which is significantly better than the baseline (\textit{p}=0.05), followed by using regions, actions, and prompts (4.29). 
However, incorporating \textit{Pano} (4.36) alongside \textit{TD+Reg+Act} (4.20) does not show a noteworthy difference.

In addition to the results above, we were also curious about the degree to which our automatic results in SPICE correlate with the human judgments. 
We measure a Kendall $\tau$ correlation between SPICE and human evaluation results of 0.6 and conclude that this is satisfactory, justifying the use of SPICE for automatic evaluation.\footnote{We also computed BLEU and ROUGE scores, however they show lower correlation with the human-assigned scores, which are omitted here.}

Our findings indicate that incorporating more information in different modalities tends to improve the performance for the generation task. 
Our semantic map abstracts information in a way that is useful for current systems, although it consists of only a single image. 
Most of our system variants that do not use panorama images performs on-par with the existing LSTM-based end-to-end approaches that use only panoramic images. 
However, the absolute performance of all models is still low, indicating that there is much room for improvement.

\subsection{Error Analysis}
\label{sec:error analysis}
Further to human evaluation score, we manually analyze the quality of the instructions generated by the same 5 system variants according to the following four aspects: 

\begin{itemize}
    \item Incorrectness: Does the prediction contain incorrect information?
    \item Hallucination: Does the prediction contain a description not corresponding to the input?
    \item Redundancy: Does the prediction contain redundant expressions and information?
    \item Linguistic problems: Is the generated instruction grammatically wrong or not fluent?
\end{itemize}

\input{tables/error_analysis}

For each experimental setting, we randomly select 15 examples.
The counts for each error type are given in Table \ref{tab:error_analysis}. 
We can see that the systems that do not use prompting or panorama images contain errors in all cases. 
Most of these errors are caused by hallucinations.
Analyzing hallucinations further, we find that the action descriptions are most prone to hallucinations, such as when left and right are confused with each other.
When regions and actions are used as input, the number of hallucinations in action descriptions goes down, but remains high in regions. 

Apart from changing the input information, 
when we train the model with prompting, the resulting instructions are less likely to include hallucinations in terms of actions and objects. 
Yet after introducing the contrastive loss, it causes redundancy and linguistic problems in the predictions. 
The language quality problems mainly consist of spelling mistakes in objects and regions, and punctuation errors when introducing the prompt and contrastive loss for training.
This may be because the contrastive loss influences the CrossEntropy loss and thus interferes with the language generation task.

%% file: figs/results_bar.tex
\begin{table}[t]
\renewcommand\arraystretch{1.3}
\resizebox{\columnwidth}{!}{%
\begin{tabular}{lcc|cc|c}
\hline
\multicolumn{1}{l}{\multirow{2}{*}{Input}} & \multirow{2}{*}{P}   & \multirow{2}{*}{C} & \multicolumn{2}{c|}{SPICE}       & Human Score   \\
\multicolumn{1}{c}{}                       &                           &                              & seen           & unseen         & unseen        \\ \hline
\multicolumn{1}{l}{TD (baseline)}          & -                         & -                            & 20.50           & 16.19           & 3.42 (5)          \\
                                           & \checkmark & -                                           & 20.79          & 15.77          & -             \\
                                           & \checkmark & \checkmark                                  & 21.78*          & 17.10           & -             \\ \hline
\multicolumn{1}{l}{TD+Reg+Act}             & -                         & -                            & 21.00             & 17.00             & 4.20 (3)          \\
                                           & \checkmark & -                                           & 21.86*          & \textbf{17.84}**  & 4.29 (2)\\
                                           & \checkmark & \checkmark                                  & 19.96          & 17.09         & 3.98 (4)         \\ \hline
\multicolumn{1}{l}{TD+Reg+Act+Pano}        & -                         & -                            & 19.87          & 17.44*          & \textbf{4.36}* (1)         \\
                                           & \checkmark & -                                           & \textbf{22.14}** & 17.79**           & -             \\
                                           & \checkmark & \checkmark                                  & 20.36          & 17.08          & -             \\ \hline
\end{tabular}
}
\caption{Automatic (SPICE) and human evaluation results with inputs of different modalities in seen and unseen environments, where P is short for prompt and C is short for contrastive loss. ** and *  indicate statistically significant difference with the baseline (p$\leq$0.01) and (p$\leq$0.05). }
\label{tab:eval_results}
\end{table}

%% file: tables/error_analysis.tex
\begin{table}[t!]
\centering
\renewcommand\arraystretch{1.5}
\resizebox{\columnwidth}{!}{%
\begin{tabular}{lcc|cccc}
\hline
\multicolumn{1}{c}{Input Information} & P & C & Incorrect & Hallucination & Redundancy & Linguistic \\ \hline
TD                              & -      & -           & 15        & 10            & 0          & 0             \\
TD + Reg + Act            & -      & -           & 15        & 10            & 0          & 1             \\
TD + Reg + Act            & \checkmark      & -           & 12        & 6             & 1          & 2             \\
TD + Reg + Act            & \checkmark      & \checkmark           & 12        & 6             & 1          & 2             \\
TD + Reg + Act + Pano & -      & -           & 11        & 6             & 0          & 0             \\ \hline
\end{tabular}%
}
\caption{Error analysis on randomly selected predictions from the systems in unseen environments, where P is short for prompt and C is short for contrastive loss. }
\label{tab:error_analysis}
\end{table}

%% file: sections/conclusion.tex
\section{Conclusion}

Our longer-term goal is to build mobile robots with spatial awareness and reasoning capabilities which can follow natural language instructions and express their intentions in natural language. 
We propose to use semantic maps  as the intermediate representation for spatial reasoning as it is a human-interpretable and light-weight approach that encodes information necessary for the navigation in a single abstract image.

In this work, we create the dataset with top-down semantic maps for R2R corpus and reframe instruction generation task as image captioning, using abstract top-down semantic map as main input. We set a baseline for the instruction generation from semantic map input. Our  experimental results show that using the top-down semantic map performs on-par with the end-to-end methods that use sequence of panorama images as input.

%% file: sections/limitations.tex
\section*{Limitations}
The current approach to the semantic map representation is missing some of the information required to generate or interpret instructions. 
For example, room names, such as {\it bathroom, bedroom,} or {\it sitting room}, are naturally used in indoor navigation instructions. However, the current single-layer semantic map representation does not encode the information about such region names. To address this in our current approach, we provide region names for each navigation point as a separate textual input.  The limitation of this approach is that it only includes the region names for the navigation points. For example, an instruction  {\it `Stop in front of the bathroom'}, the {\it bathroom} will not be included in the input because the navigation point is outside of the bathroom region. In future work, we plan to introduce a multi-layered semantic map where, in addition to encoding objects, a separate layer encodes information about regions.

Another limitation is that current semantic map encoding does not encode object properties, such as color, material, or shape. According to our analysis, object properties are mentioned in one-third of the instructions, but these would not be captured by the map. 
To address this limitation, in future work, we will encode the object properties in the semantic map. 

%% file: sections/data_code_availability.tex
\section*{Data and Code availability}
We release the code for data preparation, model training and inference, and evaluation at \url{https://github.com/chengzu-li/VLGen}, along with the prompt templates and hyper-parameter settings for experiments. 
We also release the the top-down semantic maps extracted from Habitat environment extending the existing R2R dataset, which can be obtained upon request following the guideline at \url{https://github.com/chengzu-li/VLGen}. 

%% file: appendix.tex
\section{Data Extraction}

\subsection{Conditions of Data Extraction}
\label{app:data extraction}
This section describes how the semantic map is extracted from the Habitat environment.

\paragraph{Objects}
Objects on a semantic map are represented by the bounding box with a unique color assigned to each object type. We use the $(X,Y)$ coordinates of the object's bounding box in Matterplot3D to represent them in the 2D semantic map.

There are 40 different object types labeled in the simulation environment. 
We filter out the following object types from the simulator because they are seldom mentioned in the instructions but take up a large area in the semantic map: 
\begin{verbatim}
    ['misc', 'ceiling', 'curtain', 
    'objects', 'floor', 'wall', 
    'void']
\end{verbatim}

For the buildings with multiple floors, we extract a semantic map for each floor.  Given the 3D coordinates of the object's center $(x_i, y_i, h_i)$ and the size of the object's bounding box is $(w_x, w_y, w_h)$, we use the agent's vertical position $h_{agent}$ to filter the objects for a given floor by including all objects that satisfy one of the following: 
    $$h_i - \frac{1}{2}w_h \leq h_{agent} \leq h_i + \frac{1}{2}w_h$$
    $$|h_i - h_{agent}| \leq 1.6$$

\paragraph{Regions}
For each navigation point, we determine the corresponding region by calculating whether the agent's current position is within the area of the region. 
The region's area is defined by the coordinates of the center $(x_c, y_c)$ and the sizes in width and length $(w_x, w_y)$ as a rectangle. 
We define that if the agent's location $(l_x, l_y)$ satisfies the following requirement, the region would be added into the information of this navigation point. 

$$x_c - \frac{1}{2}w_x \leq l_x \leq x_c + \frac{1}{2}w_x$$
$$y_c - \frac{1}{2}w_y \leq l_y \leq y_c + \frac{1}{2}w_y$$

\paragraph{Actions}
The actions are a closed set of \textsc{left, right, straight, stop}. 
They are determined based on the coordination of the navigation points. 
We calculate the differences in angles between the previous navigation point and the current position and define that if the differences stay within 20 degrees, the agent is heading straight. Otherwise, the agent makes a turn, with the corresponding direction depending on whether the difference is positive or negative. 

\subsection{More Information about the Extracted Data}
\label{app:quality review}
To evaluate the quality of the dataset, we randomly sample 30 examples from the training set and look into the instructions regarding the way they describe the path. 
we find that the instructions mention 2.2 landmark objects and 1.3 regions on average. 
20 out of 30 sampled instructions can be inferred simply from the top-down view.
This finding justifies the use of top-down semantic maps for navigation instruction generation to some degree. 
For the other 10 instructions, 7 out of 10 can not be inferred only from the top-down view due to the descriptive expressions about the environment.
The descriptions can be obtained from the panoramic images as supportive information (such as \textit{going upstairs} or \textit{downstairs}) and requires the interactions between different types of input. 
The other 3 instructions miss the region annotations from the simulator in R2R. 
These observations indicate that the new task we propose has the problems of weak supervision and requires the model to connect different types of inputs with each other. 

\subsection{Alignment between Colors and Objects}
Below shows the mapping between RGB pixel values and the object types (textual names) in the semantic map. 
\begin{verbatim}
[31, 119, 180], "void",
[174, 199, 232], "wall",
[255, 127, 14], "floor",
[255, 187, 120], "chair",
[44, 160, 44], "door",
[152, 223, 138], "table",
[214, 39, 40], "picture",
[255, 152, 150], "cabinet",
[148, 103, 189], "cushion",
[197, 176, 213], "window",
[140, 86, 75], "sofa",
[196, 156, 148], "bed",
[227, 119, 194], "curtain",
[247, 182, 210], "chest_of_drawers",
[127, 127, 127], "plant",
[199, 199, 199], "sink",
[188, 189, 34], "stairs",
[219, 219, 141], "ceiling",
[23, 190, 207], "toilet",
[158, 218, 229], "stool",
[57, 59, 121], "towel",
[82, 84, 163], "mirror",
[107, 110, 207], "tv_monitor",
[156, 158, 222], "shower",
[99, 121, 57], "column",
[140, 162, 82], "bathtub",
[181, 207, 107], "counter",
[206, 219, 156], "fireplace",
[140, 109, 49], "lighting",
[189, 158, 57], "beam",
[231, 186, 82], "railing",
[231, 203, 148], "shelving",
[132, 60, 57], "blinds",
[173, 73, 74], "gym_equipment",
[214, 97, 107], "seating",
[231, 150, 156], "board_panel",
[123, 65, 115], "furniture",
[165, 81, 148], "appliances",
[206, 109, 189], "clothes",
[222, 158, 214], "objects",
[255, 255, 102], "[POINT]",
[255, 255, 0], "[START]",
[255, 255, 204], "[END]",
[255, 255, 255], "[LINE]",
[0, 0, 0], "[NONNAVIGABLE]",
[150, 0, 0], "[NAVIGABLE]"
\end{verbatim}

\section{Experimental Setup}
\label{app:experimental setup}

\subsection{Hyperparameters}
We train our model for a maximum of 25 epochs using an initial learning rate of 5e-5 with linear lr scheduler. 
The batch size is set to 32 for training and 64 for validation. 
When balancing the contrastive loss and CrossEntropy loss, we assign a weight of 0.1 to the contrastive loss to make the model more focused on the generation task. 

\subsection{Data Preprocessing}
We mainly adopt the original BLIP processor for image and text inputs, but make a few modifications to the top-down semantic map. 
Because the maps are in different sizes for different rooms, we first pad them to the size of $1024\times1024$ with black pixels and apply a masking strategy to the top-down map by only selecting the nearby regions of the path. 
We keep the receptive field of the top-down view to a certain value (default to 40 pixels) within the path area. 
Then, in order to avoid introducing new pixel values when resizing, we resize the masked image with the nearest resampling interpolation strategy to $386\times386$, following the default setting of BLIP. 

\section{Prompt Design}
\label{app:prompt design}
\begin{verbatim}
prompt       : Starting from the dark 
               yellow point [objects] 
               [regions], [instruction]
For example:
[objects]    : near sofa cushion
[regions]    : in the living room region
[instruction]: exit the living room, turn 
               left, wait at the bottom 
               of the stairs. 
\end{verbatim}

\section{Experiment Results}

\subsection{Significance Test Results on SPICE Scores}
\label{app:significance test results}

We first define the indices for all of our system variants from 1 to 9 following the order of systems in Table \ref{tab:eval_results}. 
Table \ref{tab:automatic_ptest_spice} shows the full significance test on SPICE scores in unseen environments. 

\input{tables/significance_test}

\subsection{Human Evaluations}
\label{app:human eval}

\begin{figure*}[h!]
    \centering
    \includegraphics[width=1\textwidth]{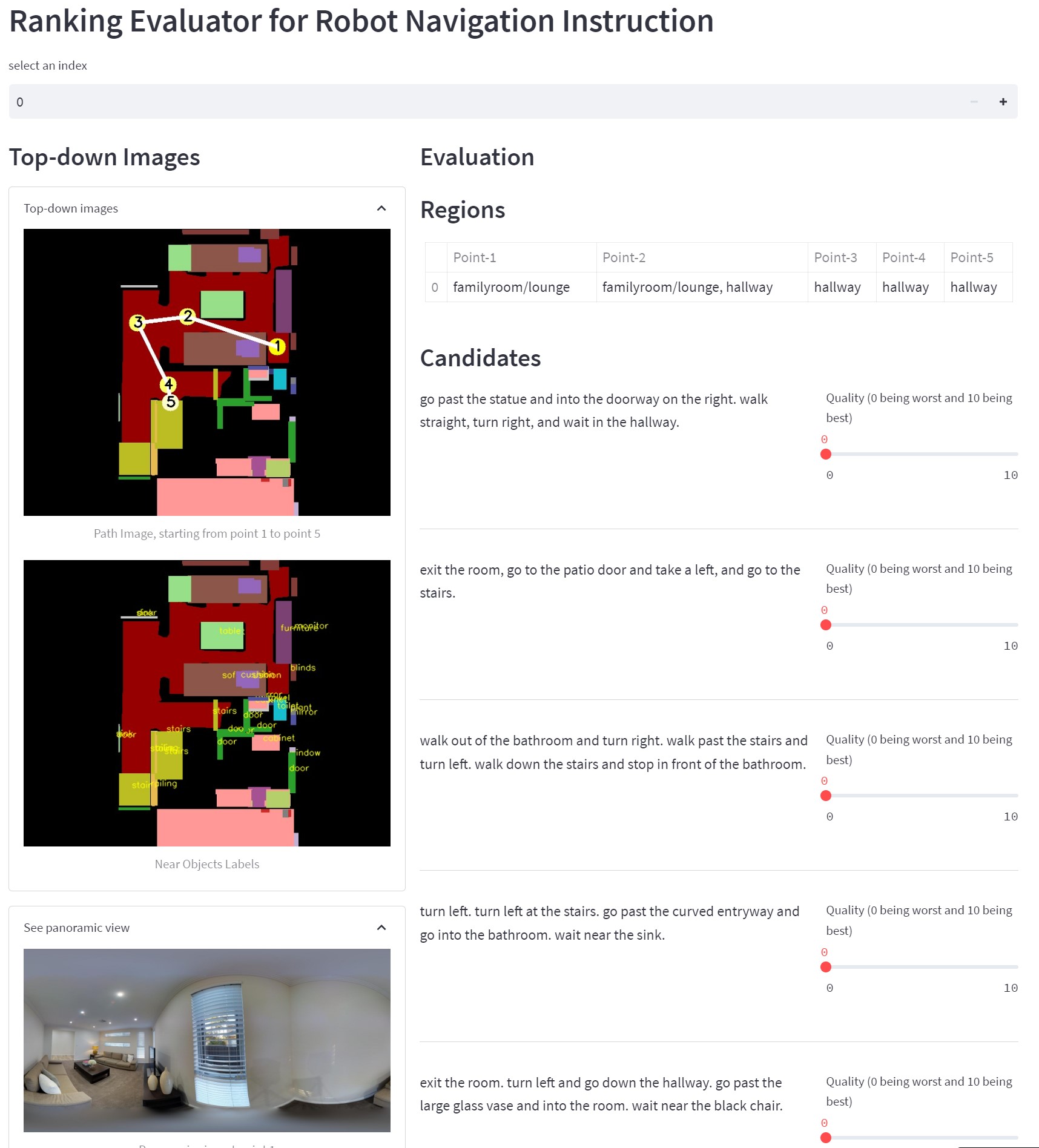}
    \caption{Screenshot of the evaluation interface for human evaluation. }
    \label{fig:eval_page}
\end{figure*}

For the evaluation page, it shows the top-down semantic map, the panoramic images and the region information as well. 
The page provides 5 generated instructions to describe the navigation path from 5 different generator models. 
The evaluator is supposed to give the quality scores for these 5 instructions based on the guidance in the instruction documentation. 
Figure \ref{fig:eval_page} shows a screenshot of the interface for human evaluation. 

\paragraph{Significance test based on human evaluation}

Table \ref{tab:sig_human} presents the two-sided permutation test results based on the human evaluation. 

\input{tables/sig_human_eval}

%% file: tables/significance_test.tex
\begin{table}[h!]
\centering
\renewcommand\arraystretch{1.5}
\resizebox{1\columnwidth}{!}{%
\begin{tabular}{ccccccccc}
                                    & \textit{2(15.77)}              & \textit{3(17.10)}              & \textit{4(17.00)}                       & \textit{5(17.84)}                       & \textit{6(17.09)}                       & \textit{7(17.44)}                       & \textit{8(17.79)}                       & \textit{9(17.08)}                       \\ \cline{2-9} 
\multicolumn{1}{c|}{\textit{1(16.19)}} & \multicolumn{1}{c|}{0.4421} & \multicolumn{1}{c|}{0.7891} & \multicolumn{1}{c|}{0.1554}          & \multicolumn{1}{c|}{\textbf{0.0030}} & \multicolumn{1}{c|}{0.1050}          & \multicolumn{1}{c|}{\textbf{0.0269}} & \multicolumn{1}{c|}{\textbf{0.0055}} & \multicolumn{1}{c|}{0.1115}          \\ \cline{2-9} 
\multicolumn{1}{c|}{\textit{2(15.77)}} & \multicolumn{1}{c|}{}       & \multicolumn{1}{c|}{0.2992} & \multicolumn{1}{c|}{\textbf{0.0310}} & \multicolumn{1}{c|}{\textbf{0.0002}} & \multicolumn{1}{c|}{\textbf{0.0175}} & \multicolumn{1}{c|}{\textbf{0.0033}} & \multicolumn{1}{c|}{\textbf{0.0004}} & \multicolumn{1}{c|}{\textbf{0.0194}} \\ \cline{2-9} 
\multicolumn{1}{c|}{\textit{3(17.10)}} & \multicolumn{1}{c|}{}       & \multicolumn{1}{c|}{}       & \multicolumn{1}{c|}{0.2432}          & \multicolumn{1}{c|}{\textbf{0.0072}} & \multicolumn{1}{c|}{0.1726}          & \multicolumn{1}{c|}{0.0505}          & \multicolumn{1}{c|}{\textbf{0.0113}} & \multicolumn{1}{c|}{0.1825}          \\ \cline{2-9} 
\multicolumn{1}{c|}{\textit{4(17.00)}} & \multicolumn{1}{c|}{}       & \multicolumn{1}{c|}{}       & \multicolumn{1}{c|}{}                & \multicolumn{1}{c|}{0.1459}          & \multicolumn{1}{c|}{0.8774}          & \multicolumn{1}{c|}{0.4536}          & \multicolumn{1}{c|}{0.1825}          & \multicolumn{1}{c|}{0.8890}          \\ \cline{2-9} 
\multicolumn{1}{c|}{\textit{5(17.84)}} & \multicolumn{1}{c|}{}       & \multicolumn{1}{c|}{}       & \multicolumn{1}{c|}{}                & \multicolumn{1}{c|}{}                & \multicolumn{1}{c|}{0.1822}          & \multicolumn{1}{c|}{0.4809}          & \multicolumn{1}{c|}{0.9318}          & \multicolumn{1}{c|}{0.1830}          \\ \cline{2-9} 
\multicolumn{1}{c|}{\textit{6(17.09)}} & \multicolumn{1}{c|}{}       & \multicolumn{1}{c|}{}       & \multicolumn{1}{c|}{}                & \multicolumn{1}{c|}{}                & \multicolumn{1}{c|}{}                & \multicolumn{1}{c|}{0.5423}          & \multicolumn{1}{c|}{0.2256}          & \multicolumn{1}{c|}{0.9913}          \\ \cline{2-9} 
\multicolumn{1}{c|}{\textit{7(17.44)}} & \multicolumn{1}{c|}{}       & \multicolumn{1}{c|}{}       & \multicolumn{1}{c|}{}                & \multicolumn{1}{c|}{}                & \multicolumn{1}{c|}{}                & \multicolumn{1}{c|}{}                & \multicolumn{1}{c|}{0.5463}          & \multicolumn{1}{c|}{0.5401}          \\ \cline{2-9} 
\multicolumn{1}{c|}{\textit{8(17.79)}} & \multicolumn{1}{c|}{}       & \multicolumn{1}{c|}{}       & \multicolumn{1}{c|}{}                & \multicolumn{1}{c|}{}                & \multicolumn{1}{c|}{}                & \multicolumn{1}{c|}{}                & \multicolumn{1}{c|}{}                & \multicolumn{1}{c|}{0.2271}          \\ \cline{2-9} 
\end{tabular}%
}
\caption{Two-sided permutation test p-values on SPICE in validation unseen environments. The row names and column names are the system indices for different systems, with the SPICE values in the parenthesis brackets. The numbers in bold are the p-values below 0.05. }
\label{tab:automatic_ptest_spice}
\end{table}

%% file: tables/sig_human_eval.tex
\begin{table}[h!]
\centering
\renewcommand\arraystretch{1.8}
\resizebox{0.5\columnwidth}{!}{%
\begin{tabular}{clccc}
                                   & \textit{4(4.20)}             & \multicolumn{1}{l}{\textit{5(4.29)}} & \multicolumn{1}{l}{\textit{6(3.98)}} & \multicolumn{1}{l}{\textit{7(4.36)}} \\ \cline{2-5} 
\multicolumn{1}{c|}{\textit{1(3.42)}} & \multicolumn{1}{c|}{0.10} & \multicolumn{1}{c|}{0.06}         & \multicolumn{1}{c|}{0.27}         & \multicolumn{1}{c|}{0.05}         \\ \cline{2-5} 
\multicolumn{1}{c|}{\textit{4(4.20)}} & \multicolumn{1}{c|}{}     & \multicolumn{1}{c|}{0.88}         & \multicolumn{1}{c|}{0.68}         & \multicolumn{1}{c|}{0.77}         \\ \cline{2-5} 
\multicolumn{1}{c|}{\textit{5(4.29)}} & \multicolumn{1}{l|}{}     & \multicolumn{1}{c|}{}             & \multicolumn{1}{c|}{0.55}         & \multicolumn{1}{c|}{0.92}         \\ \cline{2-5} 
\multicolumn{1}{c|}{\textit{6(3.98)}} & \multicolumn{1}{l|}{}     & \multicolumn{1}{l|}{}             & \multicolumn{1}{c|}{}             & \multicolumn{1}{c|}{0.47}         \\ \cline{2-5} 
\end{tabular}
}
\caption{Two-sided permutation test results between systems based on the human evaluation. The row indices and column indices are the system indices following Table \ref{tab:automatic_ptest_spice} and their quality scores in the parenthesis. }
\label{tab:sig_human}
\end{table}